\documentclass[conference]{IEEEtran}
\IEEEoverridecommandlockouts
\usepackage{cite}
\usepackage{amsmath,amssymb,amsfonts,enumerate,float,multirow,subfigure,algorithm,algorithmic}
\usepackage[usenames, dvipsnames]{color}
\usepackage{amsthm}
\usepackage{wasysym}
\usepackage{algorithm,algorithmic}
\usepackage{graphicx}
\usepackage{textcomp}
\usepackage{xcolor}

\def\BibTeX{{\rm B\kern-.05em{\sc i\kern-.025em b}\kern-.08em
    T\kern-.1667em\lower.7ex\hbox{E}\kern-.125emX}}
\begin{document}

\title{Matrix Completion with Selective Sampling\thanks{Deanna Needell was partially supported by NSF CAREER DMS \#1348721 and NSF BIGDATA DMS \#1740325}}

\author{\IEEEauthorblockN{Christian Parkinson}
\IEEEauthorblockA{\textit{Mathematics} \\
\textit{University of California, Los Angeles}\\
Los Angeles, USA\\
chparkin@math.ucla.edu}

\and 

\IEEEauthorblockN{Kevin Huynh}
\IEEEauthorblockA{\textit{Computer Science} \\
\textit{University of California, Los Angeles}\\
Los Angeles, USA\\
kevinhuynh@cs.ucla.edu}

\and

\IEEEauthorblockN{Deanna Needell}
\IEEEauthorblockA{\textit{Mathematics} \\
\textit{University of California, Los Angeles}\\
Los Angeles, USA\\
deanna@math.ucla.edu}

}

\maketitle

\begin{abstract}
Matrix completion is a classical problem in data science wherein one attempts to reconstruct a low-rank matrix while only observing some subset of the entries. Previous authors have phrased this problem as a nuclear norm minimization problem. Almost all previous work assumes no explicit structure of the matrix and uses uniform sampling to decide the observed entries. We suggest methods for selective sampling in the case where we have some knowledge about the structure of the matrix and are allowed to design the observation set. 
\end{abstract}


\section{Introduction} 

Although large-scale data is easily acquired and accessible, it is often highly incomplete. For example, data is often missing in surveys in which participants only answer a subset of questions, or sensor systems in which malfunctions or power/memory restrictions are common. Even more familiar may be the collaborative filtering problem---a problem of keen interest for companies, such as Netflix or Amazon---in which systems are tasked with recommending a subset of the vast catalogue of products to users based on sparse user histories. 


Mathematically, this is formulated as a matrix completion problem. The goal is to reconstruct a large, low-rank matrix having observed only a few entries. Let $M \in \mathbb R^{m\times n}$ be a real-valued $m \times n$ matrix and $\Omega \subset [m] \times [n]$ be a set of \emph{observed} entries. That is, we assume that we only know the entry $M_{ij}$ when the pair $(i,j)$ is in $\Omega$. From this incomplete data, we would like to reconstruct the matrix $M$. If the matrix is known to be inherently low rank, it may seem wise to look for the lowest rank representation of the observed data. That is, one may want to solve the problem \begin{equation} \min_{X} \text{rank}(X) \,\, \text{ subject to } P_{\Omega}(X) = P_{\Omega}(M),\end{equation} where $P_{\Omega}(X)_{ij} = X_{ij}$ if $(i,j) \in \Omega$ and $P_\Omega(X)_{ij} = 0$ otherwise. However, this problem is NP-hard \cite{CandPlan}, so instead Cand\'es and Recht \cite{CandRech} suggested the more tractable convex optimization problem \begin{equation} \label{eq:nucnorm} \min_{X} \| X \|_* \,\, \text{ subject to } P_{\Omega}(X) = P_{\Omega}(M),\end{equation} where $\|X \|_*$ is the nuclear norm of $X$: the sum of the singular values of $X$. In doing so, the problem is re-phrased as an $\ell_1$-minimization problem using the singular values of $X$. Since $\ell_1$-minimization lends itself to sparse solutions, solving this problem results in a low rank approximation to $M$. Several authors have proven that if the observation set $\Omega$, which is typically generated uniformly at random, is large enough, then \eqref{eq:nucnorm} leads to exact reconstruction with high probability \cite{CandPlan,CandRech,CandTao}. 

Recently, Molitor and Needell \cite{MoliNeed} adapted the ordinary nuclear norm minimization method to account for structure in the observed and unobserved entries, but most current methods for matrix completion assume little about the structure of the matrix $M$ and take the observed entries from a uniform random distribution. We propose a situation where the entries need not be observed at random, but can be chosen to account for the relationships between the columns. In application, this could be thought of as designing a survey where important questions are listed first, so that even if a user does not complete the entire survey, their answers to these questions can be used to intuit their answers to other related questions. 

\section{Selective Sampling Strategies}

Consider a scenario where $M$ is assumed to have some special structure, and $\Omega$ need not be drawn uniformly at random from $[m]\times [n]$, but can be designed. Specifically, let $\vec M_j$ for $1 \le j \le n$ denote the columns of $M$. For a set $\tau \subset [n]$ of size $t \le n$, we define $M_\tau$ to be the matrix whose columns are $\vec M_j$ for $j \in \tau$. Assume for a particular set $\tau \subset [n]$ and that the corresponding matrix $M_\tau$ has some known structure.

As a first idea, we could assume that we know the correlation matrix for $M_\tau$. However, since the map $M \mapsto \text{Corr}(M_\tau)$ is non-convex, this information is difficult to incorporate into a tractable minimization problem. Instead, if we assume that the pairwise correlations between the columns of $M_\tau$ are near $1$, then there is a strong possibility that $M_\tau$ is very low rank. Accordingly, rather than assume we have information about $\text{Corr}(M_\tau)$, we assume that we know $\text{rank}(M_\tau) = k \ll t$. This assumption is slightly stronger than assuming that the columns of $M_\tau$ are well correlated. With this assumption, we can find a basis $\{\vec v_1, \ldots, \vec v_k\}$ for the column space of $M_\tau$ along with the coordinates $B$ of the columns $\{\vec M_j\}_{j \in \tau}$ in this basis. Once we have identified these, we can use them as an additional constraint. Thus we propose the minimization problem: \begin{equation} \label{eq:selectsamp}\begin{split} \min_{X} \| X \|_{*} \,\, &\text{ subject to } P_{\Omega}(X) = P_{\Omega}(M) \\ &\text{ and } \,\, X_\tau = VB\end{split} \end{equation} where $V = [\vec v_1 \, \cdots \, \vec v_k]$. 
It remains to design a strategy for sampling entries of $M_\tau$ so that we can recover the basis $V$ and the coordinates of the columns in this basis.

\subsection{Optimal Sampling}

Assuming that $\text{rank}(M_\tau) = k$, we consider the problem of explicitly determining the relationship between the columns of $M_\tau$ while using the least possible amount of observations. That is, our goal is to find a collection of $k$ columns of $M_\tau$ (we will call this collection $M^{(k)}_\tau = [\vec M_{j_1} \, \cdots \, \vec M_{j_k}]$) and a matrix $B \in \mathbb R^{k \times (t-k)}$ such that \begin{equation} \label{eq:Mtau} M_\tau = M_{\tau}^{(k)}B.\end{equation} That is, the matrix $V$ in \eqref{eq:selectsamp} will consist of columns of $M_\tau$. 

The question is how to find $M^{(k)}_{\tau}$ and $B$ while observing as little of $M_\tau$ as possible. Notice, it suffices to extract an invertible $k\times k$ submatrix from $M_\tau$. The columns corresponding to this $k\times k$ submatrix will define $M^{(k)}_\tau$, whence we can solve for all the coefficients in $B$ with only $kt$ observations. This suggests the algorithm: 
\begin{enumerate}
\item Randomly sample $I = \{i_1,\ldots, i_k\} \subset [m]$ and $J = \{j_1,\ldots, j_k\} \subset \tau$. 
\item If the matrix $(M_{ij})_{(i,j) \in I\times J}$ is invertible, then 
\begin{enumerate} 
\item Define $M^{(k)}_\tau = [\vec M_{j_1} \, \cdots \, \vec M_{j_k}]$
\item Sample the remaining entries of the rows corresponding to $i_1,\ldots, i_k$
\item Solve for $B$ using \eqref{eq:Mtau}
\item Break loop
\end{enumerate}
\item If you reach this step, save the already observed entries and return to step 1.
\end{enumerate}

%

If $M$ is densely defined with entries coming from a continuous probability distribution, then a random $k\times k$ submatrix will almost surely be invertible, and the loop will terminate after one step (this may not be realistic with discrete data, which could result in wasting observations while looking for an invertible $k\times k$ submatrix). Counting the observed entries, step 2a will require $k^2$ observations. Determining the basis coordinates $B$ requires an additional $k(t-k)$ observations in step 2b. Then we simply need the remaining elements of the columns of $M_\tau$ to perfectly reconstruct this portion of the matrix---this requires $k(m-k)$ observations. Thus we will have observed $k(t+m-k)$ total entries; this number of observations is necessary and sufficient for perfect reconstruction of $M_\tau$, which is why we refer to this as \emph{optimal sampling}. After having used these observations, we assume that the remaining observations are taken uniformly at random from $M_{[n]\setminus \tau}$. Since we are not assuming that $M_{[n]\setminus \tau}$ has any special structure, we do not expect that there would be any advantage to selectively sampling the entries. Note, the optimization problem \eqref{eq:selectsamp} can actually be `de-coupled' at this point: simply setting $X_\tau = M^{(k)}_\tau B$ and performing nuclear norm minimization only on $X_{[n]\setminus \tau}$ which will simplify the computations. 

There are two potential ways in which we can gain accuracy using this strategy: we may gain accuracy by perfectly reconstructing $M_\tau$, and we may gain accuracy by using fewer observations while reconstructing $M_\tau$, thus saving additional observations for $M_{[n]\setminus \tau}.$ However, in application, it may not be realistic to sample entire rows or columns of the matrix.

\subsection{Finding Basis Coordinates from Random Sampling} \label{subsec:iter}

Even if $\Omega$ is constructed uniformly at random, there will likely be some invertible $k\times k$ submatrices within $M_\tau$, which can be used to intuit some relationships between the columns of $M_\tau$ without sampling full rows or columns, which may be unrealistic in practice. If we cannot sample full rows or columns, we could still attempt to find a set of bases matrices $V_\ell$, each having the same column space as $M_\tau$, and the coordinates $\vec b_\ell$ of a particular column $\vec M_\ell$ in the basis $V_\ell$, so that $\vec M_\ell = V_\ell \vec b_\ell$. This suggests the algorithm:


\begin{enumerate}
\item Set $\tau^* = \varnothing$. Repeat steps 2 - 4 until the desired amount of basis matrices $V_\ell$ and basis coordinates $\vec b_\ell$ are found.
\item Randomly sample $I = \{i_1,\ldots, i_k\} \subset [m]$ and $J = \{j_1,\ldots, j_k\} \subset \tau$. 
\item If the matrix $(M_{ij})_{(i,j) \in I\times J}$ is invertible, then 
\begin{enumerate} 
\item Choose $\ell \in \tau\setminus \tau^*$, and add $\ell$ into $\tau^*$.
\item Define $V_\ell = [\vec M_{j_1} \, \cdots \, \vec M_{j_k}].$
\item Sample the entries in column $\ell$ from each of the rows corresponding to $i_1,\ldots, i_k$. 
\item Solve for $\vec b_\ell$ using $\vec M_\ell = V_\ell \vec b_\ell$.
\item Save $V_\ell$ and $\vec b_\ell$ to use as a constraint.
\end{enumerate}
\item When you reach this step, save the already observed entries and return to step 2.
\end{enumerate} 

Having done this, we will have uncovered several relationships $\vec M_\ell = V_\ell \vec b_\ell$, and we can solve the minimization problem \begin{equation} \label{eq:selectsamp2}\begin{split}\min_{X} \| X \|_{*} \,\, & \text{ subject to } P_{\Omega}(X) = P_{\Omega}(M) \\ &\text{ and } \,\,\,\, \vec X_\ell = V_\ell \vec b_\ell, \,\, \text{ for } \ell \in \tau^* .\end{split} \end{equation} 

\begin{figure}[htbp]
\centering
\includegraphics[width=0.47\textwidth]{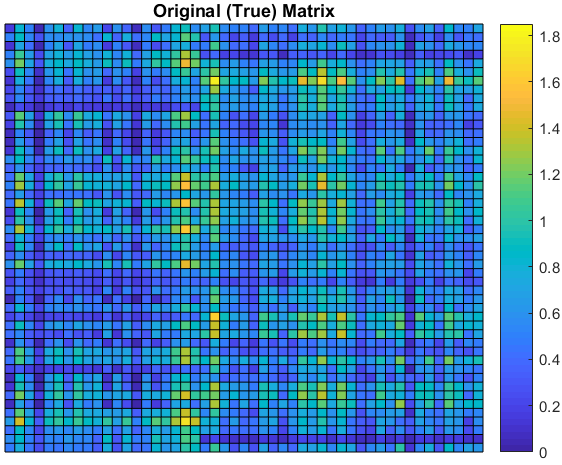}
\caption{True matrix $M \in \mathbb R^{50\times50}$, which is reconstructed in Figure~\ref{fig:pic1}. The first $t = 20$ columns have rank $k = 2$. The whole matrix has rank 6.}
\label{fig:TrueMat}
\end{figure}

\begin{figure*}
  \includegraphics[width=\textwidth]{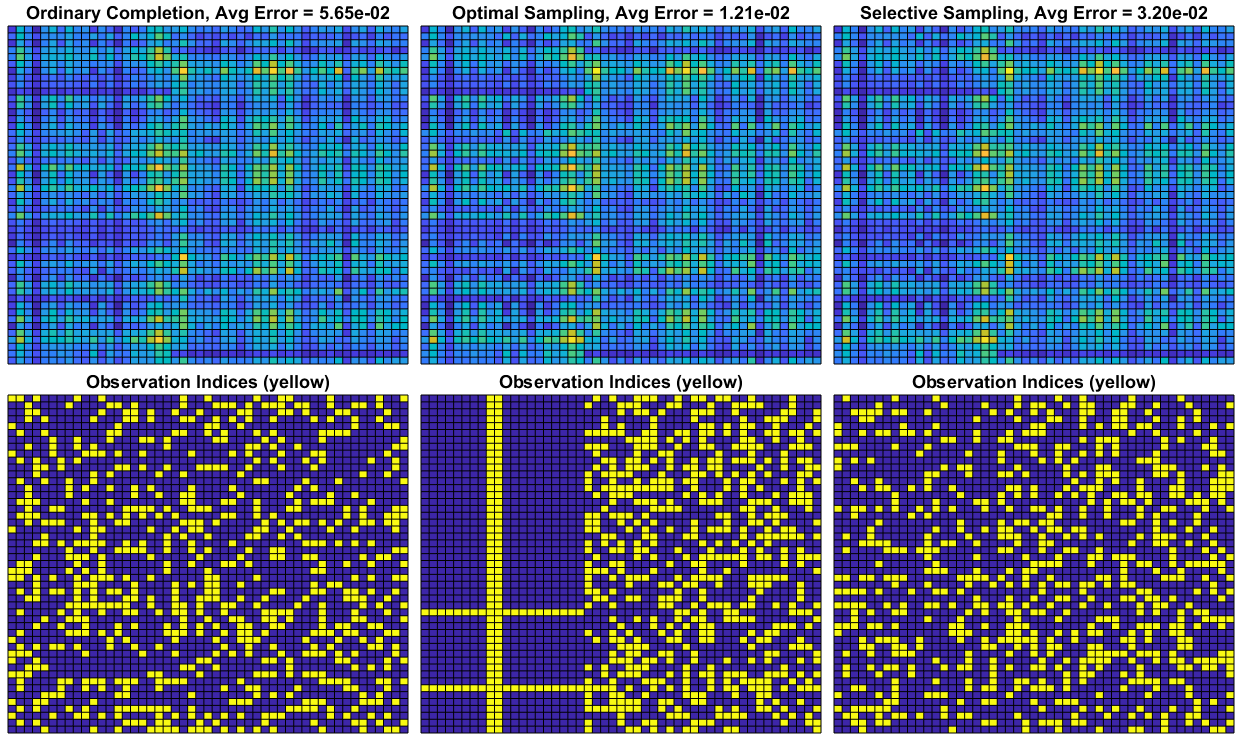}
  \caption{Comparison of reconstruction error and observation indices for different sampling strategies. Here $M_\tau$ is the first $t=20$ columns of $M$ and has rank $k = 2$. $30\%$ of entries are observed in each case.}
  \label{fig:pic1}
\end{figure*}

As we have designed it here, we are still selectively sampling the matrix, so we refer to this as \emph{selective sampling}. However, if the observations were made uniformly at random, we could search the observed entries of $M_\tau$ for invertible $k\times k$ submatrices, and perform the same steps. This formulation will not be as effective as optimal sampling, since it uses more observations and it can discover redundant relationships between the columns, but it may be more realistic in practice.

Note that in the selective sampling algorithm, we do not know the full matrix $V_\ell$ at each step. However, we do know the indices $\tau_\ell = \{j_1, \ldots, j_k \}$ which were used to construct $V_\ell$. Accordingly, in step 3e, we save all of the entries of $V_\ell$ that we know---these are the entries of $M_\tau$ which are observed. Likewise, the constraint in \eqref{eq:selectsamp2} should actually read $\vec X_\ell = X_{\tau_\ell} \vec b_\ell$ for $\ell \in \tau^*$. We are enforcing that each of these specific relationships between the columns of $X_\tau$ must hold. 

\section{Results}

We implemented the ordinary matrix completion with uniform sampling, as well as the optimal sampling method and the selective sampling method. We tested these methods on matrices $M \in \mathbb R^{50 \times 50}$ where $M_\tau$ is simply the first $t$ columns of the matrix and has rank $k < t$. We tested the methods across several different values of $t$, rank $k$ and observation rate $p$.

In Figure~\ref{fig:pic1}, we see the results of the nuclear norm minimization with uniform sampling, optimal sampling or selective sampling. Here $M$ is as described in Figure~\ref{fig:TrueMat} and the observation rate is $p = 0.3$, meaning that $30\%$ of entries are observed. The relative error is measured in the operator norm. We report the average error over $100$ trials where the observation indices are chosen independently in each trial. The 

\begin{figure}[!h]
\includegraphics[width=0.5\textwidth]{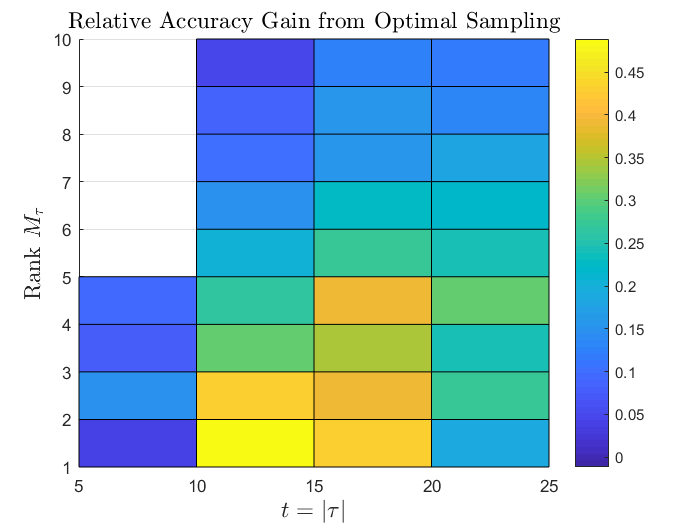}
\caption{Relative accuracy gain as a function of the size and rank of $M_\tau$. Observation rate is $p = 0.3$.}
\label{fig:accuGain}
\end{figure}

\begin{figure*}[hptb] 
\centering
\subfigure[rank$(M_\tau) = 1$]{%
\includegraphics[width=0.48\textwidth]{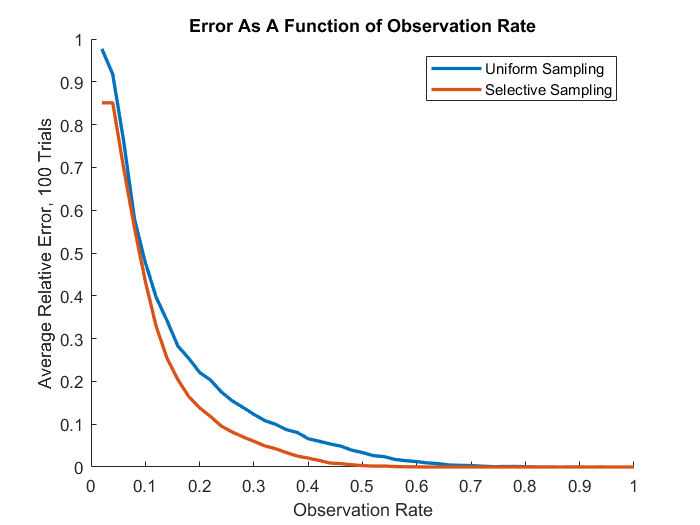}%
\label{fig:a}%
}\hfil
\subfigure[rank$(M_\tau) = 4$]{%
\includegraphics[width=0.48\textwidth]{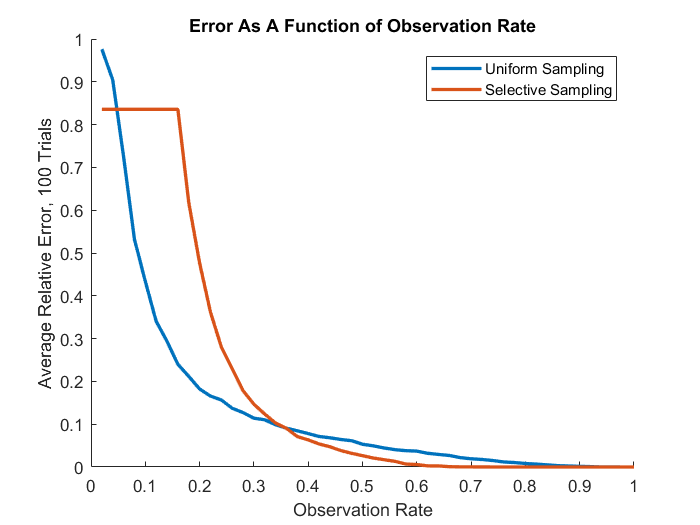}%
\label{fig:d}%
}
\label{fig:accuGain2}
\caption{Average relative error as a function of observation rate for both uniform sampling and optimal sampling and rank$(M_\tau) = 1$ or $4$.}
\end{figure*}

\noindent optimal sampling strategy led to an average accuracy gain of nearly $80\%$ and the selective sampling strategy led to an average accuracy gain of roughly $40\%$.

Next, we explored how the reconstruction errors compare when different parameters are adjusted. Recall, the optimal sampling method requires $k(t+m-k)$ observations to perfectly reconstruct $M_\tau$. We should observe accuracy gains proportional to how much smaller this number is than the expected number of observations from $M_\tau$ using uniform sampling (which is $pmt$). Thus treating $m$, the size of the matrix, as fixed, we should see the largest accuracy gains when $t$ is large, $p$ is large, or $k$ is small. First, fixing $p = 0.3$ and again working with a $50\times 50$ matrix $M$, we computed the gain in reconstruction accuracy when $k = 1, \dots ,10$ and $t = 5,10,15,20,25$. The results are displayed in Figure~\ref{fig:accuGain}. This figure aligns fairly well with our expectations. Here $M_{[n]\setminus \tau}$ has rank 4 in each case. 

Finally, we fix $t$ (the size of $M_\tau$) and vary the observation rate $p$ and the rank $k$ of $M_\tau$. The results are shown in Figure~4. In these simulations, $M$ is a $30 \times 30$ matrix and $t = 10$ so that $M_\tau$ comprises the first $1/3$ of the columns. The observation rate is allowed to vary from $p= 0$ to $p = 1$, though in the optimal sampling case, the results are not meaningful until the total number of observations $pmn$ is larger than the amount needed to construct $B$ and $M^{(k)}_\tau$, which is $k(m+t-k)$. Again, this figure aligns with our intuition. For larger $k = \text{rank}(M_\tau)$, optimal sampling requires a larger observation rate in order to see accuracy gains over uniform sampling.

\section{Conclusion}

The matrix completion problem is at the forefront of big data analysis. In application, there are often intuitive correlations between columns of the incomplete matrix: the answers to questions on a medical survey may be predictive of each other, or viewers may have similar opinions regarding movies in a given genre. Most of the previous work on this problem has focused on the general case, neglecting to consider any structure within the matrix. Building off of this work, we have suggested two methods for the matrix completion problem under the assumption that some portion of the matrix is known to be very low rank and we are allowed to design the observation set. The first method, which we termed \emph{optimal sampling}, attempts to perfectly represent the structured portion of the matrix using the minimum amount of observations. In certain scenarios, this sampling strategy led to large gains in accuracy, but it may be unrealistic in practice. Accordingly, we described a second method, \emph{selective sampling}, which forsakes perfect reconstruction of the structured portion of the matrix while still uncovering some of the structure. This method, too, led to accuracy gains in certain regimes.

\section{Acknowledgment}

To solve the minimizations problems \eqref{eq:nucnorm}, \eqref{eq:selectsamp} and \eqref{eq:selectsamp2}, we used the open source toolbox CVX, a package in MATLAB for specifying and solving convex problems \cite{CVX1,CVX2}.


\begin{thebibliography}{00}
\bibitem{CandPlan} Cand\`es, E. and Y. Plan. 2010. ``Matrix Completion With Noise.'' Proceedings of the IEEE 98 (6): 925-36. 
\bibitem{CandRech} Cand\`es, E. and B. Recht. 2009. ``Exact Matrix Completion via Convex Optimization.'' Foundations of Computational Mathematics 9 (6): 717-72. https://doi.org/10.1007/s10208-009-9045-5.
\bibitem{CandTao} Cand\`es, E. and T. Tao. 2010. ``The Power of Convex Relaxation: Near-Optimal Matrix Completion.'' IEEE Transactions on Information Theory 56 (5): 2053-80. https://doi.org/10.1109/TIT.2010.2044061.
\bibitem{CVX1} Grant, M. and S. Boyd. CVX: Matlab software for disciplined convex programming, version 2.0. http://cvxr.com/cvx, September 2013.
\bibitem{CVX2} Grant, M. and S. Boyd. Graph implementations for nonsmooth convex programs, Recent Advances in Learning and Control (a tribute to M. Vidyasagar), V. Blondel, S. Boyd, and H. Kimura (eds.), p. 95-110, Notes in Control and Information Sciences, Springer, 2008.
\bibitem{MoliNeed} Molitor, D., and D. Needell. 2018. ``Matrix Completion for Structured Observations.'' http://arxiv.org/abs/1801.09657.

\end{thebibliography}
\end{document}